%% file: main.tex
\documentclass[letterpaper, 10 pt, conference]{ieeeconf}  

\IEEEoverridecommandlockouts                              

\let\labelindent\relax
\usepackage{balance}
\usepackage[font={small}]{caption}
\usepackage{subcaption}
\usepackage{array}
\usepackage{textcomp}
\usepackage{mathtools, nccmath}
\usepackage{graphicx}
\usepackage{amsfonts}
\usepackage{amsmath}
\usepackage{amssymb}
\usepackage{algorithm}
\usepackage{algorithmic}
\usepackage{hyperref}
\usepackage{tikz}
\usepackage{arydshln}
\usepackage{multirow}
\usepackage{bm}
\usepackage{epstopdf}
\usepackage{cite}
\usepackage{siunitx}
\usepackage[inline]{enumitem}

\newcommand{\bike}[1]{{\color{red} #1}}

\newcommand{\ayush}[1]{{\color{magenta} #1}}
\newcommand{\video}[1]{{\color{magenta} #1}}

\begin{document}
\title{\LARGE \bf
Dynamic Legged Manipulation of a Ball Through \\ Multi-Contact Optimization
}
\author{Chenyu Yang, Bike Zhang, Jun Zeng, Ayush Agrawal and Koushil Sreenath
\thanks{The authors are with the Department of Mechanical Engineering, University of California, Berkeley, California, CA 94720, USA
{\tt\footnotesize \{yangcyself, bikezhang, zengjunsjtu, ayush.agrawal, koushils\}@berkeley.edu}.}
\thanks{This work was partially supported through National Science Foundation Grant CMMI-1944722.}
}
\maketitle
\begin{abstract}
The feet of robots are typically used to design locomotion strategies, such as balancing, walking, and running. However, they also have great potential to perform manipulation tasks. In this paper, we propose a model predictive control (MPC) framework for a quadrupedal robot to dynamically balance on a ball and simultaneously manipulate it to follow various trajectories such as straight lines, sinusoids, circles and in-place turning.
We numerically validate our controller on the Mini Cheetah robot using different gaits including trotting, bounding, and pronking on the ball.
\end{abstract}
\IEEEpeerreviewmaketitle
\input{introduction.tex}
\input{problem.tex}
\input{algorithm.tex}
\input{results.tex}
\input{discussion}
\input{conclusion.tex}

\balance
\bibliography{references}{}
\bibliographystyle{IEEEtranS}
\end{document}

%% file: introduction.tex
\section{Introduction}
\label{sec:intro}

\ayush{}

\subsection{Motivation}
Dynamic legged manipulation is an important strategy for humans and animals to interact with environments. 
For example, manipulation tasks like dribbling a ball, walking on stilts, and playing soccer, all require dexterous legged manipulation skills with dynamic interaction with the manipulated object \cite{peng2017deeploco, wu2009analysis, riedmiller2009reinforcement}. Not only does this repertoire extend the scope of robotic manipulation, it also paves the way to achieve agile locomotion on extremely difficult terrain \cite{zucker2010optimization}, for instance, walking on rolling boulders or toppling stepping stones. Enabling legged robots to manipulate objects using their legs shows highly dynamic motion ability and pushes the limits of robots' agility and dexterity.  

\subsection{Challenges}
Dynamic manipulation using legs places several challenges with regards to control design: 
\begin{enumerate*}
    \item The problem involves designing feedback controllers for legged robots to be able to interact with objects.
    \item The manipulated object is usually unactuated, which increases the degree-of-underactuation of the legged robot.
    \item In addition to just interacting with the object, the robot is often required to manipulate the object along a reference trajectory or to a desired pose.
    \item The manipulation of the object occurs through the locomotion of the robot which is governed by the unilateral and friction contact constraints between the robot and the object.
    \item Furthermore, the problem combines the challenges of legged locomotion, including hybrid and nonlinear dynamics with high degree-of-underactuation, as well as challenges of non-prehensile manipulation.
\end{enumerate*}

\begin{figure}[t]
    \centering
    \includegraphics[width=\columnwidth]{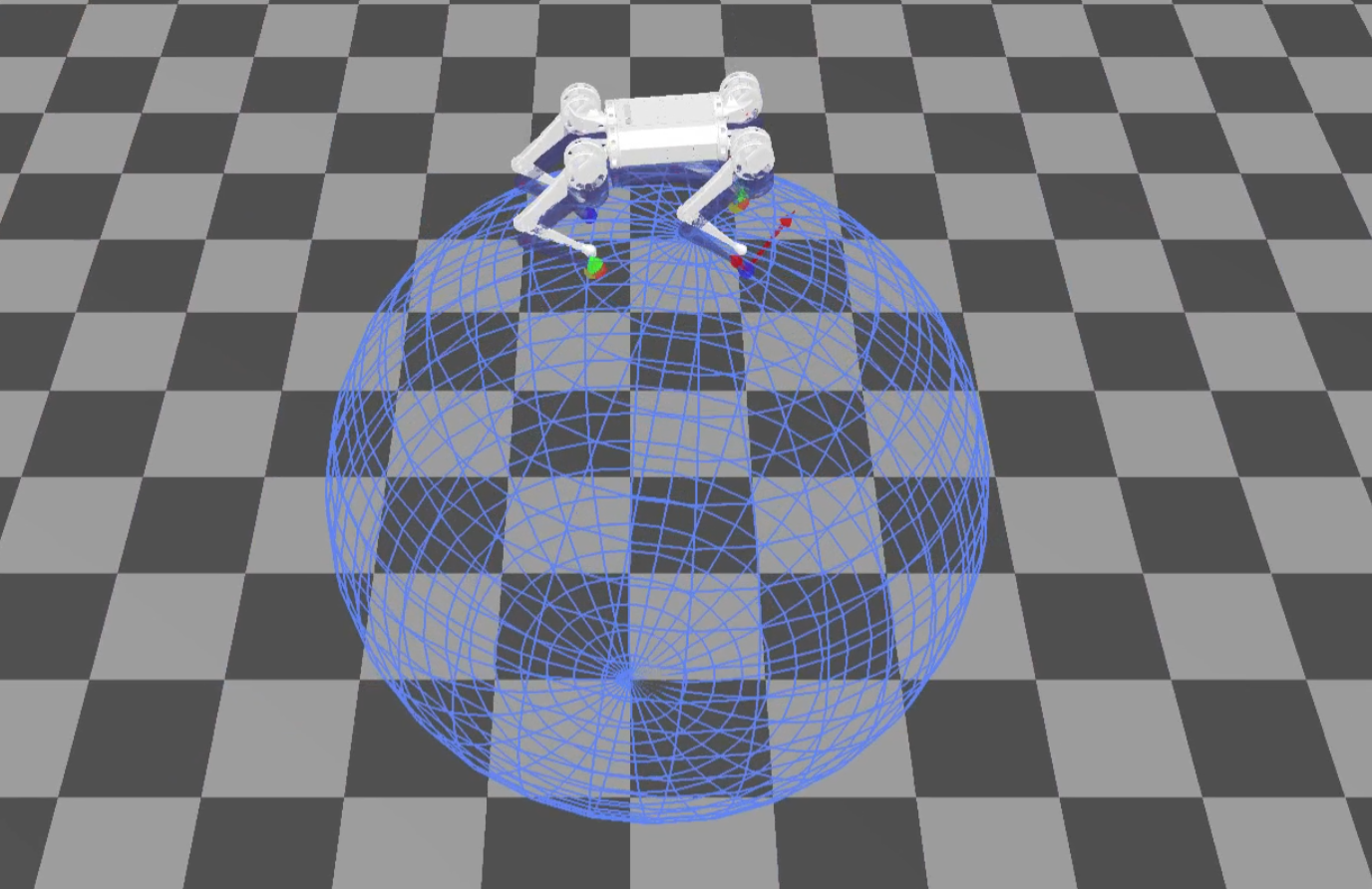}
    \caption{Simulation snapshot of Mini Cheetah dynamically balancing on and manipulating a ball. The ball has a rigid surface and the contact force between Mini Cheetah and the ball is represented by red arrows. Simulation video is at \video{\href{https://youtu.be/rIVkfudC4_8}{https://youtu.be/rIVkfudC4\_8}}. }
    \label{fig:cover}
\end{figure}

In order to address these challenges, we choose a typical scenario for analysis in this paper: a quadrupedal robot dynamically manipulating a ball to follow different trajectories while balancing on it, shown in Fig.~\ref{fig:cover}. 
A Mini Cheetah robot \cite{katz2019mini} and a rigid ball are used and the interaction between them is only through contact.

\subsection{Related Work}

There are several quadrupedal robot platforms developed for different tasks including manipulation \cite{hutter2016anymal, semini2011design, raibert2008bigdog, johnson2013legged}. These manipulation tasks are usually achieved by attaching a manipulator on a legged robot \cite{rehman2016towards, ackerman2016boston}. However, this approach does not exploit the ability of legs for manipulation tasks.


Legs of a quadrupedal robot are used for both manipulation and locomotion in \cite{wolfslag2020optimisation}, wherein the legs statically manipulate a box by holding it on both sides. In other words, two legs function as two manipulators and they do not simultaneously achieve manipulation and locomotion tasks.

A dynamic legged manipulation task has been partially implemented in \cite{zheng2011ball} for a bipedal robot balancing on and manipulating a ball. It consists of a balance controller and a footstep planner but only for 2D implementation. Our proposed method aims to provide a comprehensive control framework for 3D dynamic legged manipulation with application to Mini Cheetah manipulating a ball.

\subsection{Contribution}
The contribution of this paper is fourfold:

\subsubsection{Dynamic legged manipulation}
We formulate the legged manipulation task of a quadrupedal robot dynamically manipulating a ball in a systematic way by decoupling the whole system via contacts.

\subsubsection{Interaction model}
We develop a simplified interaction model between a quadrupedal robot and a ball based on the concept of equivalent generalized mass.

\subsubsection{Reaction-force-oriented MPC (R-MPC)}
We design a MPC strategy by taking contact forces into account to achieve the goal of dynamic ball manipulation along a given trajectory.

\subsubsection{Foot placement controller}
We present a constrained quadratic program-based foot placement controller which adapts to a spherical surface.


\subsection{Paper Structure}
This paper is organized as follows. 
In Sec. \ref{sec:problem},
we introduce the assumptions of the ball and analyze the full dynamics with the ball and the simplified dynamics.
The proposed control design is illustrated in Sec. \ref{sec:algorithm}. 
In Sec. \ref{sec:results}, we present simulation results for a Mini Cheetah robot. 
In Sec. \ref{sec:discussion}, we discuss advantages and limitations of our work, and we summarize the paper in Sec. \ref{sec:conclusion}.

%% file: problem.tex
\section{Dynamics}
\label{sec:problem}





With the goal of dynamically manipulating a ball using legs, we analyze the interaction between the robot and the ball in this section. We first make assumptions of the ball model, and then introduce the dynamical model of the robot with the ball as well as the simplified model.

We consider a rigid body model of the ball with the following assumptions: 
\begin{enumerate}
    \item The ball does not deform under the influence of external contact forces.
    \item We know the physical properties of the ball including its radius, inertia, and friction parameters.
    \item We know the ball's states including its position and velocity. 
    \item There is no slip between the ball and the foot, and the ball and the ground.
\end{enumerate}

\subsection{Dynamics of Cheetah on Ball}
\label{sec:interaction-model}

We introduce the robot's dynamical model with the consideration of the interaction with the ball. Rather than solving the dynamical equations of the ball and the robot together, as we will see, we integrate the ball's effect into the robot's model.
The dynamics of the robot and the ball can be described as follows,
\begin{align}
    \bm{A}_r  \mathbf{\ddot{q}}_{r}  + \mathbf{b}_r + \mathbf{g}_r &= \bm{\tau} + \bm{J}_r^T \mathbf{f}, \label{eq:robot-full-dynamics} \\
    \bm{A}_b \mathbf{\ddot{q}}_{b} + \mathbf{b}_b + \mathbf{g}_b &= -\bm{J}_b^T \mathbf{f}, \label{eq:ball-full-dynamics}
\end{align}
where $\mathbf{q}_{r} \in \mathbb{R}^{6+n_j}$ represents the pose of the floating base and the joints of the robot, and $n_j$ is the number of joints. $\mathbf{q}_b \in \mathbb{R}^{3}$ represents the pose of the ball, and $\bm{A}_{r/b}$, $\bm{b}_{r/b}$, $\bm{g}_{r/b}$, $\bm{J}_{r/b}$ are the generalized mass matrix, Coriolis force, gravitation force and contact Jacobian for robot or ball, respectively.
The contact force between the robot and the ball is denoted by $\mathbf{f}$, and the generalized torque of the robot actuators is represented by $\bm{\tau}$.
Here, we assume that there is no sliding between the ball and the ground, so that we can describe the ball's motion through its Euler angles as $\mathbf{q}_{b} \in \mathbb{R}^3$ with the ball's x/y position derived from these angles. Note that, with this assumption, we do not have to consider the force between the ground and the ball, and the gravitation term of the ball can be removed in \eqref{eq:ball-full-dynamics}.

For the interaction between the robot and the ball, we also assume that there is no sliding, so that we have the following constraints on the acceleration of the contact point,
\begin{equation}
    \bm{\dot{J}}_r  \mathbf{\dot{q}}_{r} + \bm{J}_r  \mathbf{\ddot{q}}_{r} = \bm{\dot{J}}_b \mathbf{\dot{q}}_{b} + \bm{J}_b \mathbf{\ddot{q}}_{b}. \label{eq:no-sliding-constraints}
\end{equation}
Substituting $\bm{J}_{b} ^{T\dagger} \bm{A}_b \mathbf{\ddot{q}}_{b} + \bm{J}_{b} ^{\bike{T}\dagger} \mathbf{b}_b + \bm{J}_{b} ^{\bike{T}\dagger} \mathbf{g}_b = -\mathbf{f}$ from \eqref{eq:ball-full-dynamics}
and $\bm{J}_{b} ^{\dagger} \bm{\dot{J}}_r \mathbf{\dot{q}}_{r} + \bm{J}_{b} ^{\dagger} \bm{J}_r  \mathbf{\ddot{q}}_{r} = \bm{J}_{b} ^{\dagger} \bm{\dot{J}}_b \mathbf{\dot{q}}_{b} + \mathbf{\ddot{q}}_{b}$ from \eqref{eq:no-sliding-constraints} into \eqref{eq:robot-full-dynamics}, we have 
\begin{equation}
\begin{aligned}
&\bm{A}_{r} \mathbf{\ddot{q}}_{r} + \mathbf{b}_{r} +\mathbf{g}_{r} +\bm{J}^T_{r} \bm{J}_{b} ^{T\dagger} \Big[ \bm{A}_{b} \bm{J}_{b} ^{\dagger} ( \bm{J}_{r} \mathbf{\ddot{q}}_{r} \\ & +\bm{\dot{J}}_{r} \mathbf{\dot{q}}_{r} - \bm{\dot{J}}_{b} \mathbf{\dot{q}}_{b} \bike{)} +\mathbf{b}_{b} + \mathbf{g}_{b} \Big] = \bm{\tau} ,
\end{aligned}
\label{eq:equivalent-dynamics-derive}
\end{equation}
where the $\bm{J}_{b} ^{\dagger}$ is a pseudo inverse of $\bm{J}_b$.

Ignoring the Coriolis force and gravitation terms of the ball $\mathbf{b}_b$ and $\mathbf{g}_b$, and the terms involving the derivatives of Jacobians,
we get the equivalent dynamics based on \eqref{eq:equivalent-dynamics-derive} as follows,
\begin{equation}
    \bm{\tilde{A}}  \mathbf{\ddot{q}}_{r} + \mathbf{b}_r + \mathbf{g}_r =  \bike{\bm{\tau}}, \label{eq:equivalent-dynamics}
\end{equation}
where $\bm{\tilde{A}}$ represents the \emph{equivalent generalized mass} as follows. 
\begin{equation}
    \bm{\tilde{A}} = \bm{A}_r + \bm{J}_r^T \bm{J}_b^{T\dagger} \bm{A}_b \bm{J}_b^{\dagger} \bm{J}_r.
\end{equation}
We use this equivalent generalized mass to describe the dynamics and generate torques in the whole body impulse control (WBIC) which will be introduced in Sec.~\ref{sec:MIT-WBIC}.

\subsection{Simplified Model}
\label{sec:simplified-Model}
Having presented the full dynamics model with consideration of the ball, we now present a simplified model of the robot that will be used in the reaction-force-oriented model predictive control (R-MPC) in Sec. \ref{sec:mpc-contact}.
As introduced in \cite{di2018dynamic}, we use the lumped mass model,
\begin{align}
    m \mathbf{\ddot{p}}_r &= \sum_{i=1}^{n_c} \mathbf{f}_i + \mathbf{g}, \\
    \dfrac{d}{dt} (\bm{I \omega}_r) &= \sum_{i=1}^{n_c} \mathbf{r}_i \times \mathbf{f}_i,
\end{align}
where $\mathbf{p}_r$, $\bm{\omega}_r$, and $\mathbf{g}$ are three dimensional vectors denoting the robot's position, angular velocity, and acceleration due to gravity, all in the global frame. The mass and the moment of inertia of the robot are denoted by $m$ and $\bm{I}$ respectively. $n_c$ is the number of contacts, and $\mathbf{r}_i$, $\mathbf{f}_i$ are the relative position to the center-of-mass and the contact force of the $i$-th foot, respectively.

\begin{figure*}[t]
    \centering
    \includegraphics[width=\textwidth]{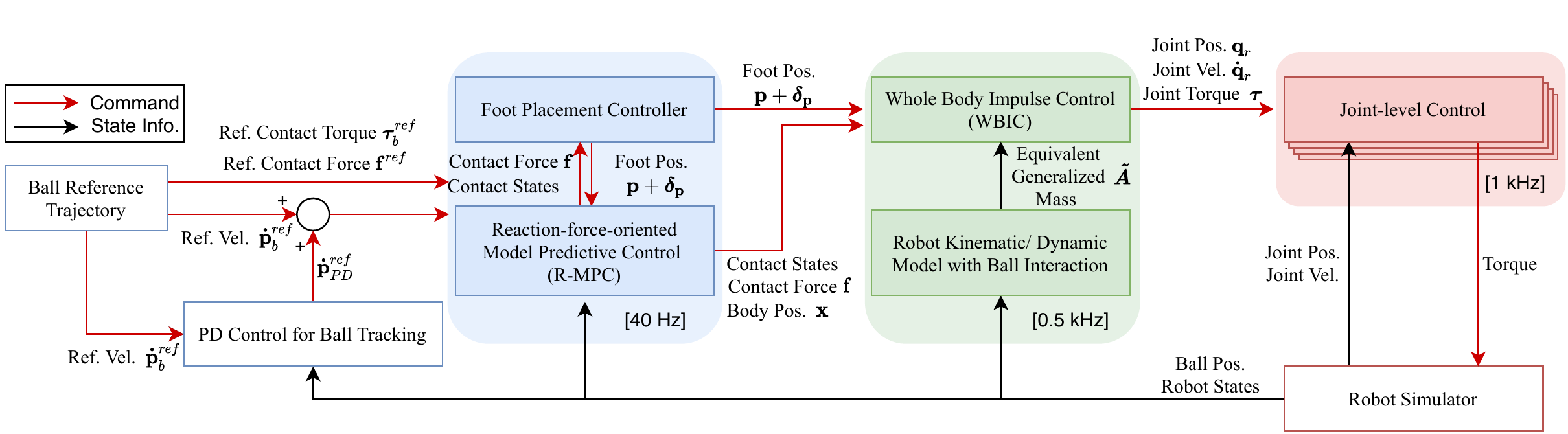}
    \caption{Control framework for Mini-Cheetah manipulating a ball along a given ball trajectory. Firstly gait types, reference velocity, contact forces and the torque to the ball could be given by user or calculated from ball's reference trajectory. The reaction-force-oriented MPC computes reference contact forces and foot/body position commands, and this allows us to use WBIC to compute joint torque, position, and velocity commands, which is sent to each joint controller. The foot placement controller is added to adjust the planned foot placement, optimizing alternatively with R-MPC. We also consider the interaction between the robot and the ball in the dynamics for WBIC.
    }
    \label{fig:algorithm-pipeline}
\end{figure*}

During the stance phase, as in \cite{di2018dynamic}, the MPC makes three assumptions: 
1) roll and pitch angles are small,
2) states are close to the reference trajectory,
3) roll and pitch velocities are small and off-diagonal terms of the inertia tensors are small. From \cite{di2018dynamic}, the linearized discrete-time dynamics of the system is then expressed as follows,
\begin{equation}\label{eq:linear_robot_model}
    \mathbf{x}(k+1) = \bm{A}_k \mathbf{x}(k) + \bm{B}_k \mathbf{f}(k) + \hat{\mathbf{g}},
\end{equation}
where $\mathbf{x}$ represents body configurations and velocities,
$\mathbf{f}(k)$ and $\hat{\mathbf{g}}$ represent the contact forces from the ball and the robot's gravity. These variables are given by, 
\begin{align}
\mathbf{x}&=
[\mathbf{\Theta}_r^T \quad \mathbf{p}_r^{T} \quad \mathbf{\omega}_r^{T} \quad \dot{\mathbf{p}}_r^{T}] ^{T},\\
    \mathbf{f} &= [\mathbf{f}_1 \quad \mathbf{f}_2 \quad ... \quad \mathbf{f}_{n_c}]^T, \label{eq:contact_force}\\
    \hat{\mathbf{g}} &= [\mathbf{0}_{1 \times 3} \quad \mathbf{0}_{1 \times 3} \quad \mathbf{0}_{1 \times 3} \quad \mathbf{g}^T]^T,
\end{align}
where $\mathbf{\Theta}_r$ is the body orientation. The discrete-time dynamics matrices $\bm{A}_k$ and $\bm{B}_k$ are defined as follows,
\begin{align}
    \bm{A}_k &= \begin{bmatrix}
    \mathbf{1}_{3 \times 3} & \mathbf{0}_{3 \times 3} & \bm{R}_z \Delta t & \mathbf{0}_{3 \times 3} \\
    \mathbf{0}_{3 \times 3} & \mathbf{1}_{3 \times 3} & \mathbf{0}_{3 \times 3} & \mathbf{1}_{3 \times 3} \Delta t \\
    \mathbf{0}_{3 \times 3} & \mathbf{0}_{3 \times 3} & \mathbf{1}_{3 \times 3} & \mathbf{0}_{3 \times 3} \\
    \mathbf{0}_{3 \times 3} & \mathbf{0}_{3 \times 3} & \mathbf{0}_{3 \times 3} & \mathbf{1}_{3 \times 3}
    \end{bmatrix}, \label{eq:mpc-A-matrix} \\
    \bm{B}_k &= \begin{bmatrix}
    \mathbf{0}_{3 \times 3} & ... & \mathbf{0}_{3 \times 3} \\
    \mathbf{0}_{3 \times 3} & ... & \mathbf{0}_{3 \times 3} \\
    _{\mathcal{G}} I^{-1}[\mathbf{r}_1]_{\times}\Delta t  & ... & _{\mathcal{G}} I^{-1}[\mathbf{r}_n]_{\times}\Delta t \\
    \mathbf{1}_{3 \times 3} \Delta t / m & ... & \mathbf{1}_{3 \times 3} \Delta t / m
    \end{bmatrix}, \label{eq:mpc-B-matrix}
\end{align}
where $_{\mathcal{G}} I$ is the inertia matrix with respect to the global frame, and $\bm{R}_z$ is the matrix of the body rotation around z-axis.  
$[\mathrm{x}]_{\times} \in so(3)$ 
is defined as the skew-symmetric matrix for cross products. 
This discrete-time dynamics is then used in the R-MPC in Sec. \ref{sec:mpc-contact}.

%% file: algorithm.tex
\section{Control Design}
\label{sec:algorithm}


Having presented the assumptions of the ball as well as the dynamics, we now proceed to present our control design for dynamic legged manipulation.

\subsection{Control Framework}

Our proposed control framework extends the control hierarchy in \cite{kim2019highly} by taking two key issues of ball manipulation into account: 
\begin{enumerate*}
\item how to model the quadruped robot balancing on the ball and
\item how to make use of the contact force and position to drive the ball.
\end{enumerate*}
The first issue was addressed in Sec.~\ref{sec:interaction-model}, where we analyzed the interaction model and introduced the \emph{equivalent generalized mass}.
For the second issue, we propose a \emph{reaction-force-oriented MPC} (R-MPC) and a \emph{foot placement controller}. 
The R-MPC is designed to follow the reference state and contact force, see Sec. \ref{sec:mpc-contact}. 
The foot placement controller first plans the foot placement according to the Raibert heuristic from \cite{kim2019highly}, and then adjusts the foot placement point to generate a reference torque to the ball, see Sec. \ref{sec:foot-placement-compensator}. 
Lastly, the ball reference trajectory and PD control for ball tracking, whose commands are used for R-MPC and the foot placement controller, are described in Sec. \ref{sec:Ball-trajectory-speed-profile-and-PD}.

Our control framework is described in Fig.~\ref{fig:algorithm-pipeline}. The ball reference trajectory calculates reference velocity, torque and force of the ball according to different scenarios. 
A PD controller is added to alter the velocity command, which takes the relative position and velocity of ball and robot into account.
The foot placement controller is at the same level with reaction-force-oriented MPC, sharing the commands and the state information. They work together to track the commands of the reference robot state and the reference contact torque and force to the ball. 
R-MPC and the foot placement controller optimize the foot placement and the contact force profile alternatively. 
The WBIC tracks the foot position command and the contact force from R-MPC and the foot placement controller using the interaction model. The joint-level control executes commands from WBIC and controls the motors.

\subsection{Reaction-force-oriented MPC}
\label{sec:mpc-contact}
After introducing the control framework, 
we next present the reaction-force-oriented MPC (R-MPC), which plans the contact force with the simplified dynamics from Sec.~\ref{sec:simplified-Model} using the reference robot trajectory and the reference contact force.
The R-MPC minimizes the tracking error and the deviation of contact forces, under the friction cone constraints. It is a constrained quadratic programming (QP) problem, and its formulation is described as follows, 

\noindent\rule{\columnwidth}{0.4pt}
\textbf{Reaction-force-oriented MPC (R-MPC)}:
\begin{equation}
\begin{aligned}
\min_{\mathbf{x}, \mathbf{{f}}} & \sum_{k=0}^m ||\mathbf{x}(k+1) - \mathbf{x}_r^{ref}(k+1)||_{\bm{Q}} + ||\mathbf{f}(k) - \mathbf{f}^{ref}(k)||_{\bm{R}} \\
\text{s.t.} \quad 
& \mathbf{x}(k+1)=\bm{A}_{k} \mathbf{x}(k)+\bm{B}_{k} \mathbf{f}(k)+\hat{\mathbf{g}},\\
& |\mathbf{f}_i^x(k)| \leq \mu \mathbf{f}^z(x) \quad i=1\dots n_c,\\
  & |\mathbf{f}_i^y(k)| \leq \mu \mathbf{f}^z(k) \quad i=1\dots n_c , \\ 
  & \mathbf{f}_i^z(k)\geq 0 \quad ~~~~~~~~i=1\dots n_c.
\end{aligned}
\label{eq:Reaction-force-oriented-MPC}
\end{equation}
\noindent\rule{\columnwidth}{0.4pt}
Here $\bm{Q}$ and $\bm{R}$ are positive definite weight matrices. The friction cone constraint is simplified to a four-side pyramid constraining the x/y direction of the force of each $i$-th contact $\mathbf{f}_i^x(k)$, $\mathbf{f}_i^y(k)$. 
The R-MPC problem optimizes robot state $\mathbf{x}$ and contact force $\mathbf{f}$ that appears as the input to the simplified system in \eqref{eq:linear_robot_model}. 
Different from \cite{di2018dynamic}, we consider the reference contact forces regulated from the ball's reference trajectory in \eqref{eq:contact-force-pd}, as we hope to control the robot and the ball at the same time. 
We compute the reference contact force through a PD control from the ball's reference trajectory. 

The reference trajectory $\mathbf{x}_r^{ref}$ is similar with \cite{di2018dynamic}. 
The reference x/y position of the robot are determined by integrating the reference velocities $\mathbf{\dot{p}}_{r}^{ref}$. The reference yaw and the yaw rate of the robot is from the commanded direction \cite{kim2019highly}. The $z$ position of the robot is a user defined constant. The other states (roll, pitch, roll rate, pitch rate, and z velocity) are set to 0. 

The calculation of the reference velocities $\mathbf{\dot{p}}_{r}^{ref}$ and the reference contact forces $\mathbf{f}^{ref}(k)$ will be described in Sec.~\ref{sec:Ball-trajectory-speed-profile-and-PD}.
Besides being passed to WBIC, the solution of $\mathbf{f}(k)$ is then used as inputs to the foot placement controller, which will be described next in Sec. \ref{sec:foot-placement-compensator}. The output of the foot placement controller affects the relative position of planned foot placement $\mathbf{r}_i$, which consequently affects the dynamics matrix $\bm{B}_k$ in the next control iteration.





\subsection{Foot Placement Controller}
\label{sec:foot-placement-compensator}

The foot placement contributes to the 
torque generated on the ball, so we plan to exploit this potential advantage.
In order to find a foot placement that yields desired torque on the ball, the foot placement should ideally be a decision variable in the R-MPC. 
However, this makes the optimization problem nonconvex, which is computationally expensive to implement in real-time.

As a trade-off between the convexity of the optimization problem and the achievement of the control objective, we formulate the R-MPC and foot placement controller as two separate optimization problems with the results of one being used by the other.
Specifically, the R-MPC is solved while holding the foot placement as a constant. 
Then, the foot placement controller is executed to generate the reference foot placement, and the new foot placement will be used in the R-MPC in the next control iteration.

\begin{figure}
    \centering
    \includegraphics[width=0.95\linewidth]{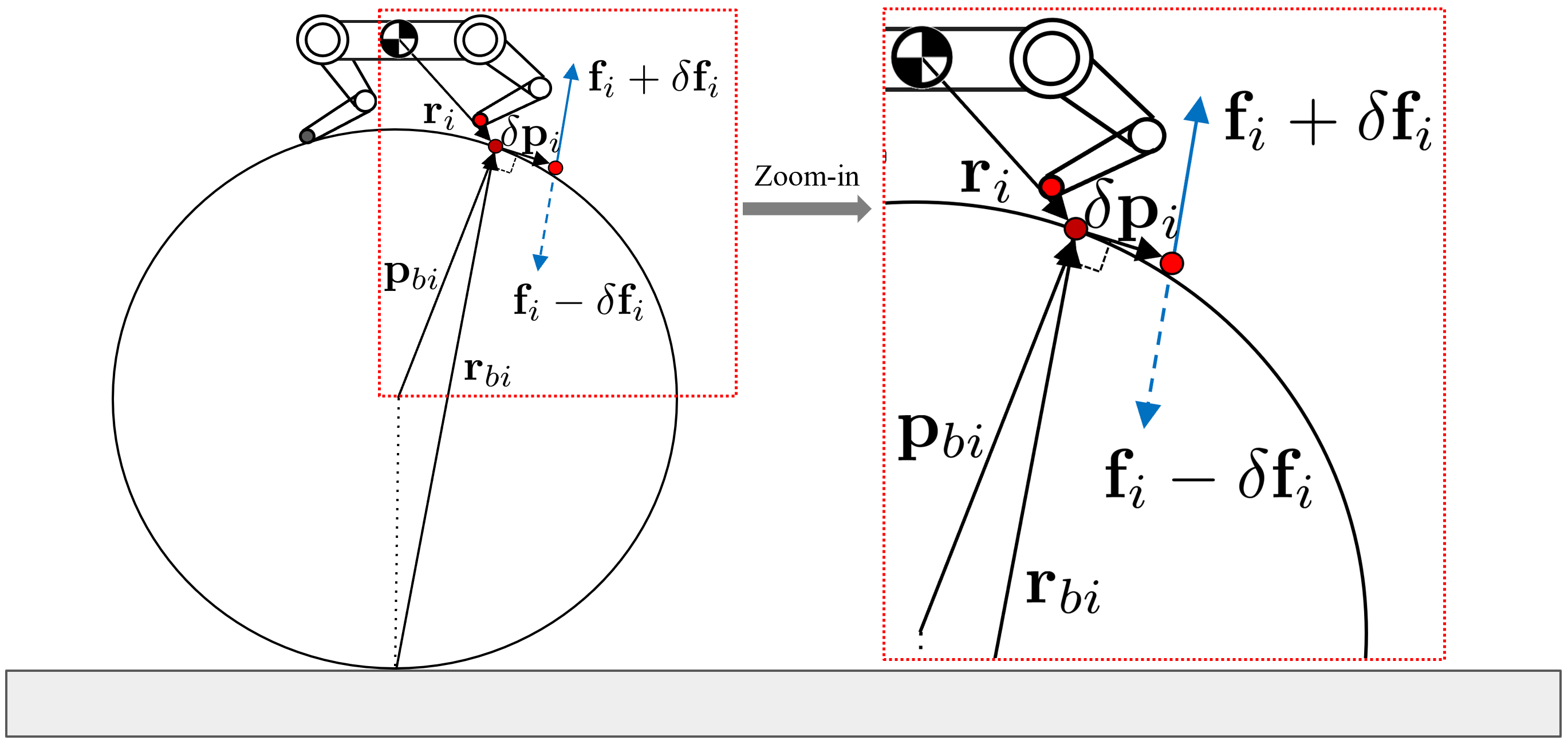}
    \caption{Mini-Cheetah on the ball and its zoomed view. The foot placement controller adjusts the swing foot placement after R-MPC by calculating a displacement $\delta \mathbf{p}_i$, so that the torque of the contact force at the adjusted foot placement $(\mathbf{r}_{bi} + \delta \mathbf{p}_i) \times ( - \mathbf{f}_i-\delta \mathbf{f}_i)$ meets the commanded ball torque.}
    \label{fig:foot-placement-compensator}
\end{figure}

The foot placement controller only changes the planned foot placement of the swing foot. 
Once a foot is lifted off and becomes a swing foot, its placement is set according to the Raibert heuristic from \cite{kim2019highly} that is presented as follows,
\begin{equation}
\mathbf{r}_{i}^{\mathrm{cmd}}=\mathbf{p}_{\mathrm{shoulder}, i}+\mathbf{p}_{\mathrm{symmetry}}+\mathbf{p}_{\mathrm{centrifugal}},
\end{equation}
where,
\begin{equation}
\begin{split}
\mathbf{p}_{\text{shoulder},i}&=\mathbf{p}_{r}+\boldsymbol{R}_{z}\left(\psi_{k}\right) \mathbf{l}_{i}, \\
\mathbf{p}_{\text{symmetry}}&=\frac{t_{\text {stance }}}{2} \mathbf{\dot{p}}_r+k\left(\mathbf{\dot{p}}_r-\mathbf{\dot{p}}_r^{\mathrm{ref}}\right), \\
\mathbf{p}_{\text{centrifugal}}&=\frac{1}{2} \sqrt{\frac{h}{g} }\mathbf{\dot{p}_r} \times \omega_r^{\mathrm{ref}}.
\end{split}
\label{eq:raibert-explain}
\end{equation}
In \eqref{eq:raibert-explain}, 
$\mathbf{l}_{i}$ is $i$-th leg shoulder location with respect to the body frame, and $h$ is the height of the CoM.
$\mathbf{p}_{\text{symmetry}}$ is called Raibert heuristic that 
uses foot placements to stabilize the horizontal CoM dynamics, $t_{\text{stance}}$ is the stance duration of the current gait cycle, $\omega_r^{\text{ref}}$ and $\mathbf{\dot{p}}_r^{\mathrm{ref}}$ are the reference robot angular and linear velocities respectively.

After R-MPC plans the contact force profile, the foot placement controller alters the foot placement using the following optimization program,


\noindent\rule{\columnwidth}{0.4pt}
\textbf{Foot Placement Controller:}
\begin{equation}
  \begin{aligned}
\min_{\boldsymbol{\delta}_{\mathbf{p}}, \boldsymbol{\delta}_{\mathbf{f}}} &  \left\|\boldsymbol{\delta}_{ \bm{\tau}{r}}(\boldsymbol{\delta}_{ \mathbf{p}} ,\boldsymbol{\delta}_{ \mathbf{f}}) \right\|_{Q_{r}} + \left\|\bm{\tau}^{ref}_{b}(\boldsymbol{\delta}_{ \mathbf{p}} , \boldsymbol{\delta}_{\mathbf{f}}) -\bm{\tau}_{b}(\boldsymbol{\delta}_{\mathbf{p}} , \boldsymbol{\delta}_{\mathbf{f}}) \right\|_{Q_{b}}\\
   + & || \boldsymbol{\delta}_{\mathbf{p}}||_{{\bm{R}}_{\boldsymbol{\delta}{\mathbf{p}}}} + ||\boldsymbol{\delta}_{ \mathbf{f}}||_{{\bm{R}}_{ \boldsymbol{\delta}{\mathbf{f}}}} \\
  \text{s.t.} \ & \boldsymbol{\delta}_{\mathbf{p}i} \times \mathbf{p}_{bi} = 0.
  \end{aligned}
\label{eq:qp-based-compensator}
\end{equation}
\noindent\rule{\columnwidth}{0.4pt}
Here, $Q_r$, $Q_b$, $R_{\boldsymbol{\delta}{\mathbf{p}}}$, and $R_{\boldsymbol{\delta}{\mathbf{f}}}$ are positive definite weight matrices, $ \boldsymbol{\delta}_{\mathbf{p}}, \boldsymbol{\delta}_{\mathbf{f}}$ are concatenated vectors of the deviations of contact positions $\boldsymbol{\delta}_{\mathbf{p}i}$ and contact forces $\boldsymbol{\delta}_{\mathbf{f}i}$ of each foot. By ignoring the higher order terms of infinitesimal variations of positions and forces $\boldsymbol{\delta}_{\mathbf{p}
i}$ and $\boldsymbol{\delta}_{\mathbf{f}i}$, we can express the change of the contact torque to robot $\boldsymbol{\delta}_{\bm{\tau}r}$, and the contact torque to the ball $\bm{\tau}_b$ as follows,

\begin{equation}
\boldsymbol{\delta}_{\bm{\tau}{r}}(\boldsymbol{\delta} _{\mathbf{p}} , \boldsymbol{\delta}_{\mathbf{f}}) = \sum_{i=1}^{4} \boldsymbol{\delta}_{\mathbf{p}i} \times \mathbf{f}_i + \mathbf{r}_i \times \boldsymbol{\delta}_{ \mathbf{f}i},
\end{equation}
\begin{equation}
  \bm{\tau}_{b}( \boldsymbol{\delta}_{\mathbf{p}} , \boldsymbol{\delta}_{\mathbf{f}})  = \sum_{i=1}^{4}  \mathbf{r}_{bi} \times (-\mathbf{f}_i) + \boldsymbol{\delta}_{\mathbf{p}i} \times (-\mathbf{f}_i) +  \mathbf{r}_{bi} \times (-\boldsymbol{\delta}_{\mathbf{f}i}) .
\end{equation}
Here, $\mathbf{r}_i$ and $\mathbf{r}_{bi}$ are the relative position from the robot CoM to foot placement and the relative position from ball's ground contact point to the foot placement, respectively. 
$\mathbf{p}_{bi}$ is the relative position from the ball's center to the foot contact point. The contact force $\mathbf{f}_i$ comes from R-MPC.
We illustrate the geometric relation in Fig. \ref{fig:foot-placement-compensator}.

The foot placement controller is a multi-objective quadratic program where the first term minimizes the resulting torque change on the robot, and the second term tracks the reference torque to be applied on the ball. 
The third and fourth terms ensure that the solution is in the proximal neighbor of the original point. 
The optimization constraint guarantees that the optimized foot placement is on the surface of the ball.
Note that $\boldsymbol{\delta}_\mathbf{f}$ is introduced as a slack variable and it is not used anywhere else. 


\subsection{Whole-body Impulse Control}
\label{sec:MIT-WBIC}
Following the reference from the foot placement controller and the R-MPC, the WBIC tries to incorporate both body posture stabilization and contact force execution with a full dynamics model of the robot. 
Apart from the use of the equivalent generalized mass $\tilde{\bm{A}}$, we did not make any other modification to the WBIC described in \cite{kim2019highly}. The WBIC calculates joint level commands from the reference contact force and body trajectory in the following steps.
First, an acceleration command $\ddot{\mathbf{q}}_r^{\mathrm{cmd}}$ of the robot's configuration is computed to execute a set of user-specified prioritized tasks. Then, the following QP is solved,
\begin{equation}
\begin{aligned}
& \min_{\boldsymbol{\delta}_{\mathbf{f}}, \boldsymbol{\delta}_{q}} \boldsymbol{\delta}_{\mathbf{f}}^{\top} \boldsymbol{Q}_{1} \boldsymbol{\delta}_{\mathbf{f}}+\boldsymbol{\delta}_{q}^{\top} \boldsymbol{Q}_{2} \boldsymbol{\delta}_{q}\\
\text{s.t.} \quad&
\boldsymbol{S}_{f}(\bm{\tilde{A}} \ddot{\mathbf{q}}_r+\mathbf{b}_r+\mathbf{g}_r)=\boldsymbol{S}_{f} \boldsymbol{J}_{r}^{\top} \mathbf{f}  ~~~~~ \text{(floating base dyn.)} \\
& \ddot{\mathbf{q}}_r=\ddot{\mathbf{q}}_r^{\mathrm{cmd}}+\left[\begin{array}{c}\boldsymbol{\delta}_{q} \\ \mathbf{0}_{n_{j}}\end{array}\right]  ~~~~~~~~~~~~~~~~~~~~~~ \text{(acceleration)} && ~ \\ 
& \mathbf{f}=\mathbf{f}^{\text{R-MPC}}+\boldsymbol{\delta}_{\mathbf{f}} ~~~~~~~~~~~~~~~~~~~~~~~~~~ \text{(reaction forces)} && ~\\
& \boldsymbol{W} \mathbf{f} \geq \mathbf{0} ~.~~~~~~~~~~~~~~~~~~~~~~ \text{(contact force constraints)} && ~
\end{aligned}
\end{equation}
Here, $\boldsymbol{\delta}_{\mathbf{f}}$ and $\boldsymbol{\delta}_q$ are relaxation variables for the reaction forces and the floating base acceleration.
$\bm{\tilde{A}},\mathbf{b}_r,\mathbf{g}_r,\boldsymbol{J}_{r}$ 
are defined in Sec.~\ref{sec:interaction-model}.
The $\boldsymbol{Q}, \boldsymbol{S}_{f}, \boldsymbol{W}$ are weight matrices of the deviation, the row selection matrix of the floating base, and the matrix of the friction cone and the normal direction of the contact surface. The solution $\boldsymbol{\delta}_{\mathbf{f}}$ is used to adjust the planned contact force ${\mathbf{f}}$ to satisfy the full dynamics. At last, the torque commands can be computed by plugging the contact forces and the acceleration into the robot's dynamics and passed to the joint level controllers.

\subsection{Ball Reference Trajectory and Tracking}
\label{sec:Ball-trajectory-speed-profile-and-PD}

The ball reference trajectory generates the velocity command of the robot as well as the reference contact force and the reference torque to be applied on the ball. Note that the yaw angle of the ball is not considered in the ball reference trajectory.


The reference x/y velocities of the robot, $\mathbf{\dot{p}}_r^{ref}$, are obtained from the reference x/y velocities of the ball $\mathbf{\dot{p}}_b^{ref}$ and a feedback term, $\mathbf{\dot{p}}_{PD}^{ref}$,
\begin{equation}
    \mathbf{\dot{p}}_{r}^{ref}=\mathbf{\dot{p}}_{b}^{ref}+\mathbf{\dot{p}}_{PD}^{ref}.
    \label{eq:robot-reference-velocity}
\end{equation}
\noindent
Here, $\mathbf{\dot{p}}_{PD}^{ref}$ is a feedback component to minimize the tracking error between the instantaneous ball and the robot positions,
\begin{equation}
    \mathbf{\dot{p}}^{ref}_{\text{PD}} = - \bm{k}_p^{\mathbf{v}}(\mathbf{p}_r - \mathbf{p}_b) - \bm{k}_d^{\mathbf{v}} \mathbf{\dot{p}}_r.
\end{equation}
where $\mathbf{p}_b$ and $\mathbf{p}_r$ are the position of the ball and the robot. $\mathbf{\dot{p}}_r$ is the velocity of the robot's CoM.

We compute the reference contact force $\mathbf{f}^{ref}$ and the reference torque  $\bm{\tau}^{ref}_{b}$ through a PD control from ball's reference trajectory,
\begin{equation}
    \mathbf{f}^{ref}(k) = -\bm{k}_p^{\mathbf{f}} \mathbf{e}_p -\bm{k}_d^{\mathbf{f}} \mathbf{e}_v, \label{eq:contact-force-pd}
\end{equation}
\begin{equation}
    \bm{\tau}^{ref}_{b} = \left[\begin{array}c 0 \\ 0\\ 1  \end{array}\right] {\times} (-\bm{k}_p^{ \bm{\tau}}  \mathbf{e}_p -\bm{k}_d^{ \bm{\tau}} \mathbf{e}_v ),
    \label{eq:tau-ref-pd}
\end{equation}
where $\mathbf{e}_p=\mathbf{p}^{ref}_b - \mathbf{p}_b$ and $\mathbf{e}_v = \mathbf{\dot{p}}^{ref}_b - \mathbf{\dot{p}}_b$ represent ball's position and velocity error. $\bm{k}_p^{\mathbf{f}}$, $\bm{k}_d^{\mathbf{f}}$, $\bm{k}_p^{ \bm{\tau}}$, and $\bm{k}_d^{ \bm{\tau}}$ are PD gains.



%% file: results.tex
\section{Results}
\label{sec:results}

\begin{figure}
    \centering
    \begin{subfigure}{0.95\linewidth}
       \includegraphics[width=0.98\linewidth]{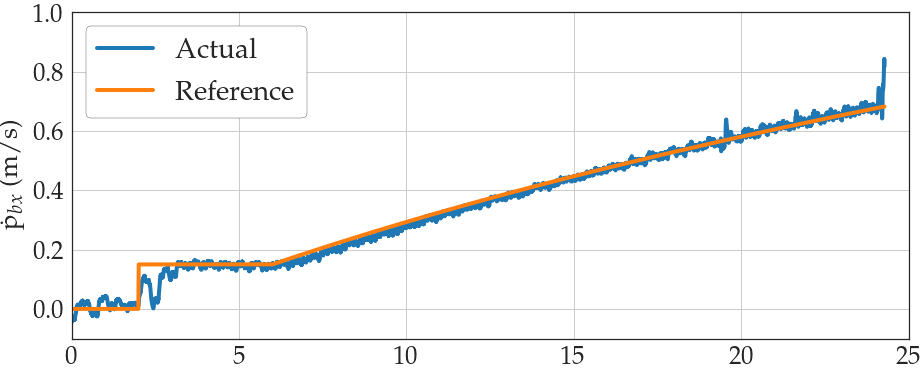}
       \caption{Proposed Controller}
       \label{fig:highspeed_proposed}
    \end{subfigure}
    \begin{subfigure}{0.95\linewidth}
       \includegraphics[width=0.98\linewidth]{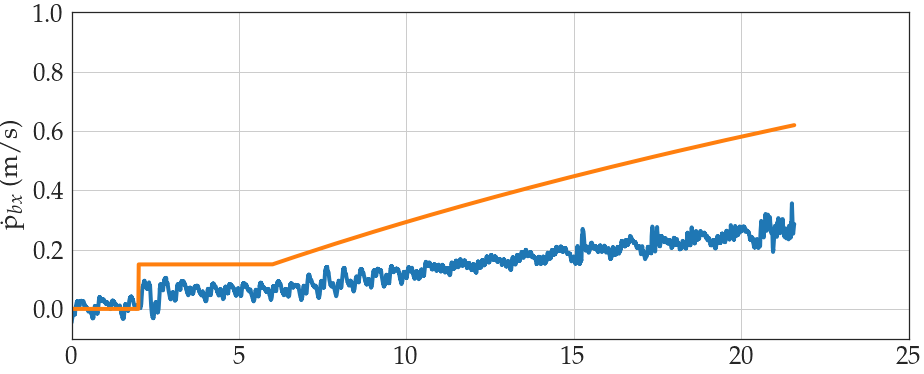}
       \caption{Baseline Controller}
       \label{fig:highspeed_baseline}
    \end{subfigure}
    \caption{The comparison of reference speed tracking between (a) our proposed controller and (b) the baseline controller. 
    The blue line represents the actual speed, which tries to catch up with an increasing command shown as the orange line.}
    \label{fig:Hspeed-plot}
\end{figure}

\begin{figure}
    \centering
    \begin{subfigure}{0.95\linewidth}
       \includegraphics[width=0.98\linewidth]{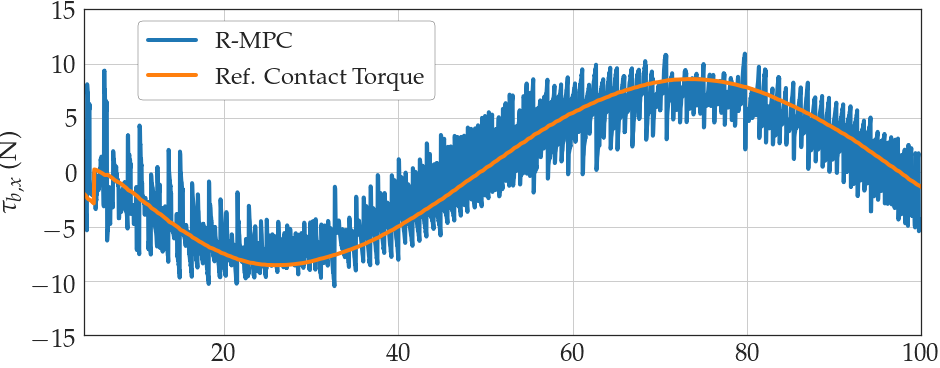}
       \caption{Proposed Controller}
       \label{fig:ball_torque_proposed}
    \end{subfigure}
    \begin{subfigure}{0.95\linewidth}
       \includegraphics[width=0.98\linewidth]{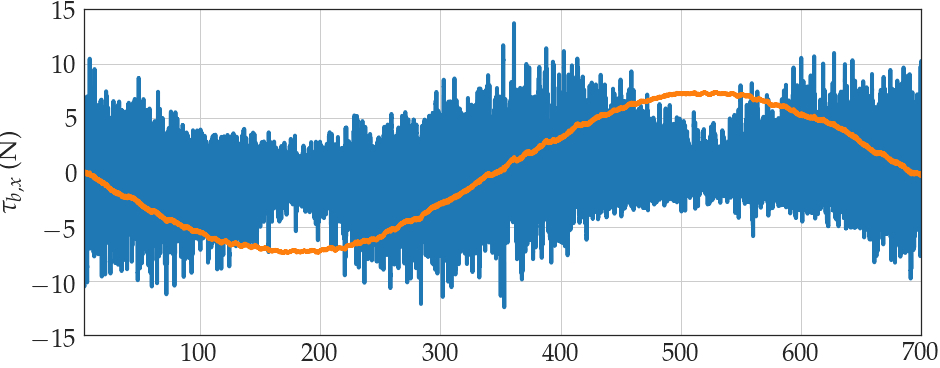}
       \caption{Baseline Controller}
       \label{fig:ball_torque_baseline}
    \end{subfigure}
    \caption{The comparison of reference ball torque command tracking between (a) our proposed controller and (b) the baseline controller. The blue line shows the torque that can be exerted on the ball from the R-MPC, which is a cross product of current relative contact position and the contact force. The orange line is the ball's torque command from the ball reference trajectory.
    }
    \label{fig:BTau-plot}
\end{figure}

Having introduced our control design for dynamic legged manipulation, we next present simulation results that validate the control strategy in this section.

\subsection{Simulation Setup}


The simulation environment is set up using the MIT Cheetah Software\footnote{\url{https://github.com/mit-biomimetics/Cheetah-Software}}, and a compliant contact model is utilized to compute interaction forces between the robot and the ball based on Featherstone's algorithm \cite{featherstone}. The friction coefficient $\mu$ is set as $0.9$. The ball's radius is 1m. 
We use the controller from \cite{kim2019highly} that is designed to walk on flat ground as our \emph{baseline controller}. 
Specifically, the baseline controller uses the control framework in Fig.~\ref{fig:algorithm-pipeline} without the foot placement controller, with Robot Kinematic/Dynamic Model of solely the robot, and original MPC instead of R-MPC.
Note that the PD Control for Ball Tracking module is included and tuned separately in both our proposed controller and the baseline controller.
Also, both the proposed controller and the baseline controller have adapted the direction of gait to the normal direction at the contact point. The ball is created from blender \cite{blender} with 2562 points. 


\subsection{Performance Evaluation}

\begin{figure}
    \centering
    \begin{subfigure}{0.95\linewidth}
       \includegraphics[width=0.98\linewidth]{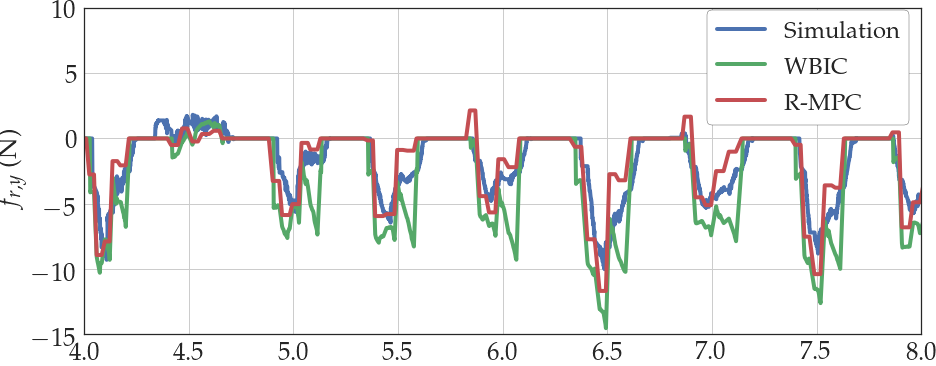}
       \caption{Proposed Controller}
       \label{fig:Reaction_force_proposed}
    \end{subfigure}
    \begin{subfigure}{0.95\linewidth}
       \includegraphics[width=0.98\linewidth]{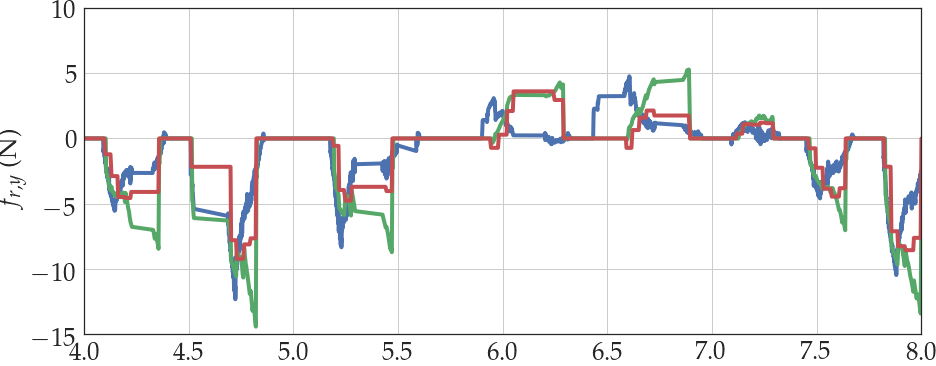}
       \caption{Baseline Controller}
       \label{fig:Reaction_force_baseline}
    \end{subfigure}
    \caption{The comparison of the contact force tracking in y-direction between (a) our proposed controller and (b) a baseline controller. Curves shown in the figure are the planned contact force from R-MPC (red), the contact force solved in WBIC (green), and the actual contact force in simulation (blue)}
    \label{fig:RF-plot}
\end{figure}


The speed tracking performance is shown in Fig.~\ref{fig:Hspeed-plot}. 
We set the reference velocity in ball reference trajectory to be a step followed by a ramp, and the yaw rate to be zero. 
Our proposed controller can track the reference and continuously accelerate until 0.75 m/s. 
The baseline controller can also track a step input but with more tracking errors. 
Under mild acceleration, the baseline controller maintains stability as well. However, it can not get close to the 0.75m/s speed limit, 
and it has a steady state error for tracking a constant acceleration.
The proposed control strategy improves the speed tracking performance and extends the range of speed for manipulating a ball.

Fig.~\ref{fig:BTau-plot} shows the effectiveness of our proposed control design for reference ball torque command tracking. 
The robot manipulates a ball to follow a circular trajectory in this scenario.
The orange line is the ball's torque command from the ball reference trajectory, as shown in \eqref{eq:tau-ref-pd}. The blue line shows the torque that can be exerted on the ball from the R-MPC and the foot placement controller, which is a cross product of current relative contact position and the contact force. While the proposed controller keeps tracking the reference commands, the baseline controller loses tracking and is more noisy.


\begin{figure*}
    \centering
    \begin{subfigure}{0.24\linewidth}
       \includegraphics[width=0.98\linewidth]{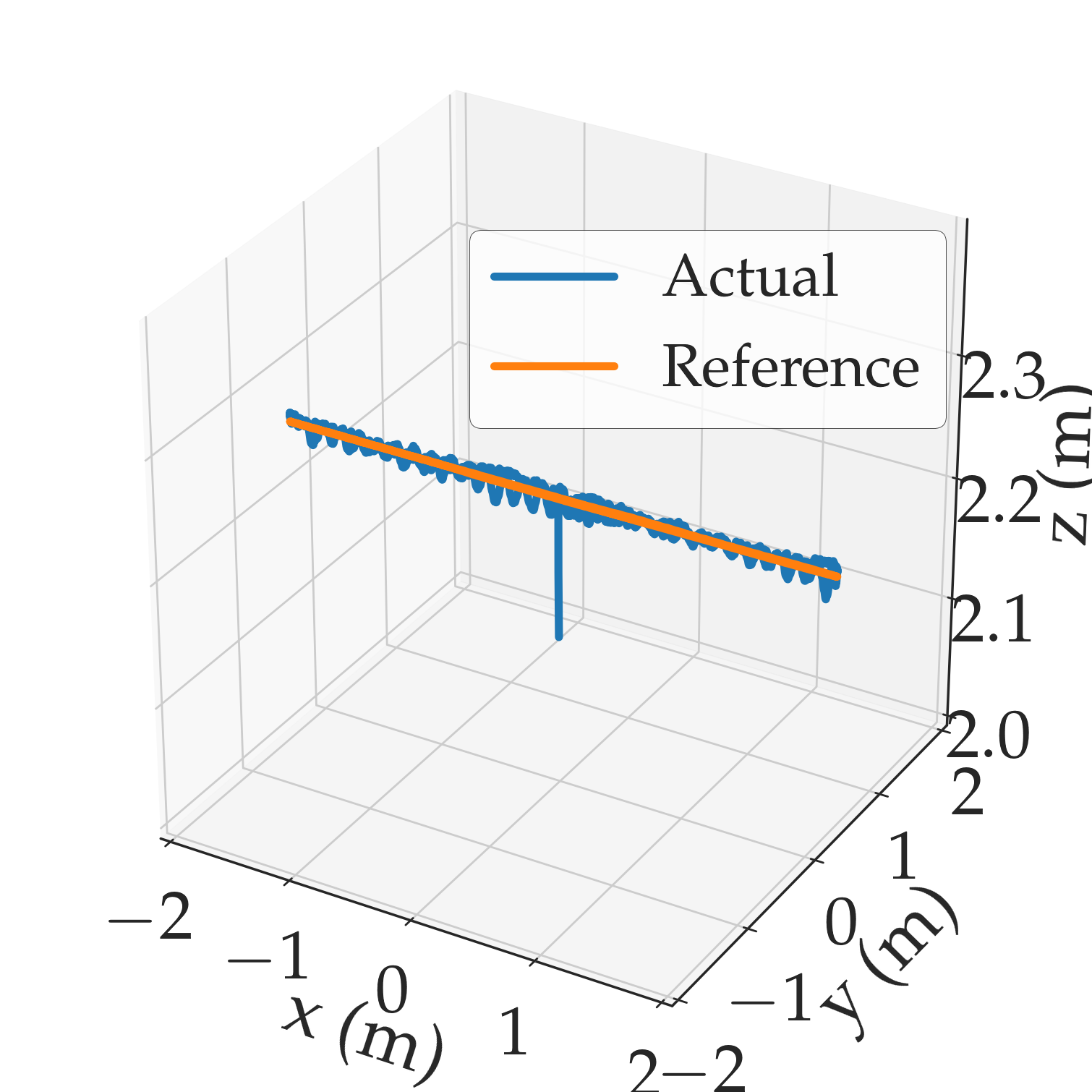}
       \caption{Tracking a straight line}
       \label{fig:scenario-line}
    \end{subfigure}
    \begin{subfigure}{0.24\linewidth}
       \includegraphics[width=0.98\linewidth]{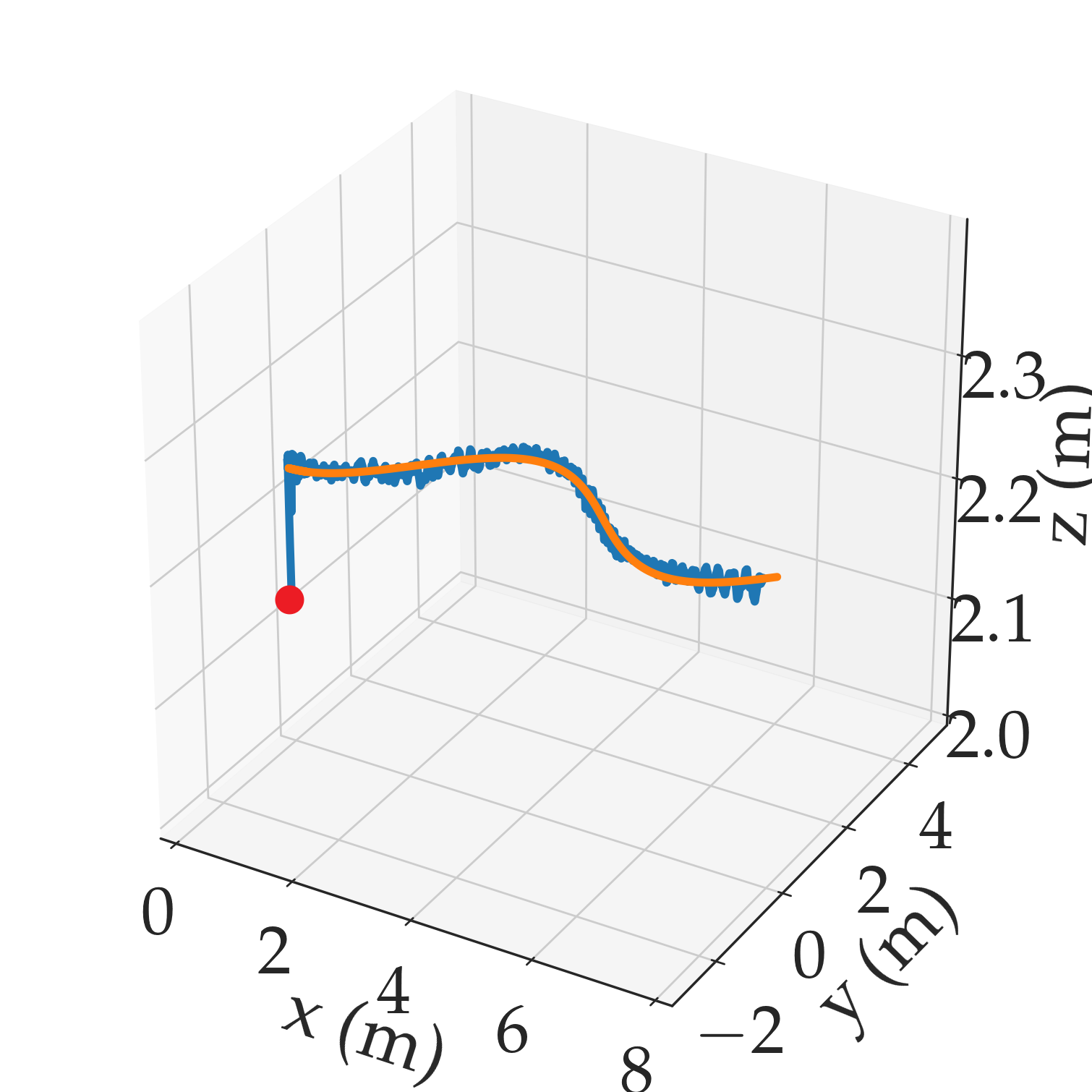}
       \caption{Tracking a sinusoid curve}
       \label{fig:scenario-sin}
    \end{subfigure}
    \begin{subfigure}{0.24\linewidth}
       \includegraphics[width=0.98\linewidth]{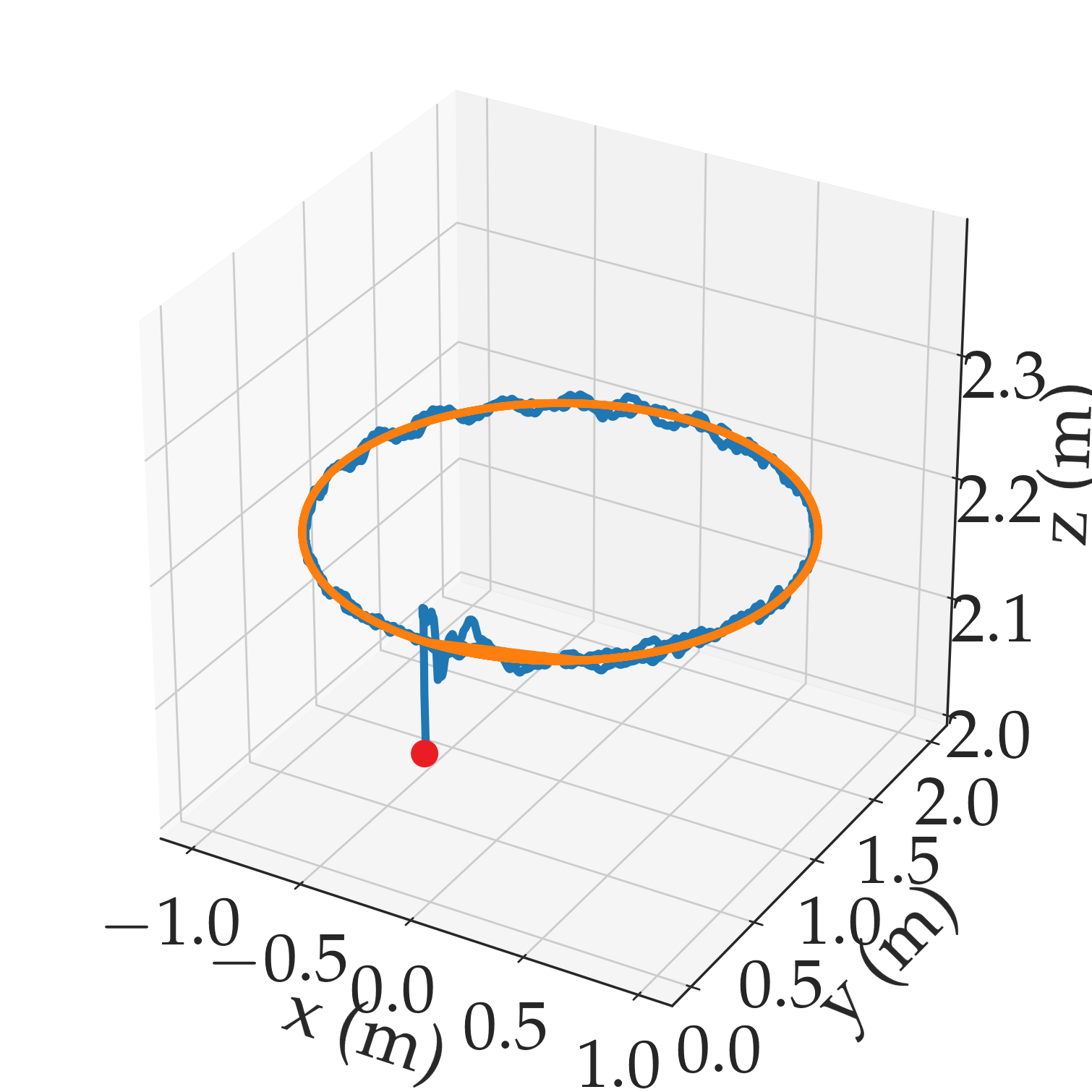}
       \caption{Tracking for a circle}
       \label{fig:scenario-circle}
    \end{subfigure}
    \begin{subfigure}{0.24\linewidth}
       \includegraphics[width=0.98\linewidth]{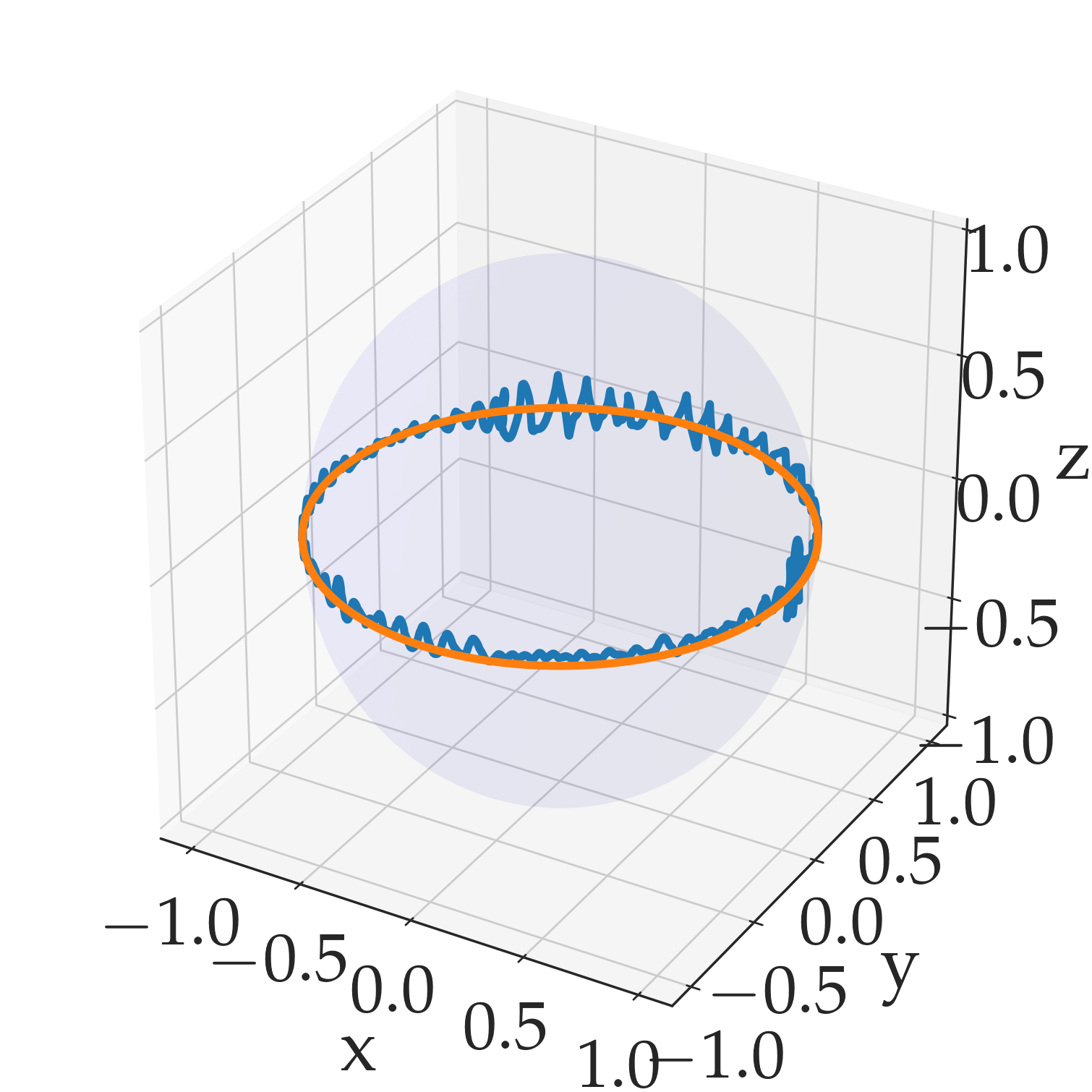}
       \caption{Yaw angle tracking}
       \label{fig:scenario-spin}
    \end{subfigure}
    \caption{The 3D tracking performance of Mini Cheetah in different scenarios: (a) tracking a straight line (b) tracking a sinusoid curve (c) tracking a circle (d) orientation tracking. The orange lines are reference trajectories and blue lines are actual trajectories. Red dots indicate the initial position of the robot. The orientation of the robot is shown as point on an unit sphere in (d).}
    \label{fig:scenario}
\end{figure*}


Fig.~\ref{fig:RF-plot} evaluates how MPC and WBIC track the contact force command in the same scenario of circular trajectory tracking. 
We only include data from the front left leg, and other legs are similar.
In the control hierarchy, the contact force from WBIC should follow the command of R-MPC (proposed controller) or MPC (baseline controller), and by performing the torque commands from WBIC, the real contact force in simulation is at best to be the same as WBIC expected. The closer real contact force to the WBIC's command suggests a more accurate dynamic model. 
Using our proposed controller with equivalent generalized mass, the contact force in simulation follows the command of WBIC. 
However, the contact forces of the baseline controller have larger errors in both timing and scale.
Using the proposed controller, the mean squared error (MSE) between the real contact force and the WBIC command among four legs of the first $10$ seconds is $1810.2$, smaller than that of the baseline controller as $2109.4$.
Note that the commands from R-MPC/MPC are also different, as the baseline controller can not stabilize the robot well.

\subsection{Different Scenarios}

Besides comparing the performance with the baseline controller, we implement our controller for different scenarios to further validate our control design.

\subsubsection{Different Gaits}
The proposed controller is tested using different gaits. The experimental \href{https://youtu.be/rIVkfudC4_8}{video} shows that trotting, bounding and pronking gaits all work for a Mini Cheetah robot dynamically balancing on a ball. 

\subsubsection{Different Reference Trajectories}
Fig.~\ref{fig:scenario} shows the performance of our proposed control design in four different scenarios. In Fig.~\ref{fig:scenario-line} - \ref{fig:scenario-circle}, Mini Cheetah manipulates a 2 kg ball to follow a straight line, a sinusoid, and a circle at velocity 0.3 m/s. 
Note that the reference yaw angle is the angle of the robot, rather than the ball.
The reference position and velocity in \eqref{eq:contact-force-pd} and \eqref{eq:tau-ref-pd} are generated from the ball reference trajectory. 
For example, for the circular trajectory tracking scenario, the ball reference trajectory is given as, 
\begin{equation}
\left[\begin{array}c {p}_{bx}^{ref} \\{p}_{by}^{ref}  \\ {\dot{p}}_{bx}^{ref} \\ {\dot{p}}_{by}^{ref} 
\end{array}\right]  
= 
\left[\begin{array}c r \sin (t v^{ref}/r ) \\
- r \cos (t v^{ref}/r )\\
v^{ref} \cos (t v^{ref}/r )\\
v^{ref} \sin (t v^{ref}/r )
\end{array}\right] ,
\end{equation}
and the robot yaw angle,
\begin{equation}
    \psi_r^{ref}  = tv^{ref} /r,
\end{equation}
where $r=0.7$ m is the radius of the circular trajectory, $v^{ref}=0.3$ m/s is the reference velocity, and $t$ is the current time. 
The reference ball velocity $\mathbf{\dot{p}}_b^{ref}$ in \eqref{eq:robot-reference-velocity} is the reference velocity $\dot{p}_{bx/y}^{ref}$ plus a feedback tracking of reference ball position $p_{bx/y}^{ref}$.
Notice that the Mini Cheetah starts with position errors from the origin. Fig.~\ref{fig:scenario-spin} shows the orientation of Mini Cheetah 
following the yaw angle command, which is illustrated by points on a unit sphere. From these plots, we can clearly see that Mini Cheetah tracks these predefined trajectories well. 


Fig.~\ref{fig:ScenarioCircle} shows the detailed tracking errors when Mini Cheetah manipulates a ball to follow a circle: roll, pitch, yaw angle, $\dot x$, $\dot y$, $\dot z$ of Mini Cheetah. 
We observe that there is a slight tracking delay on tracking yaw angle and robot velocities. In order to make the robot move forward, we need some pitch angle errors to provide torque on the ball and there is a steady state tracking error around $0.2$ rad on the pitch angle.

%% file: discussion.tex
\section{Discussion}
\label{sec:discussion}

\begin{figure*}
    \centering
    \includegraphics[width=0.95\linewidth]{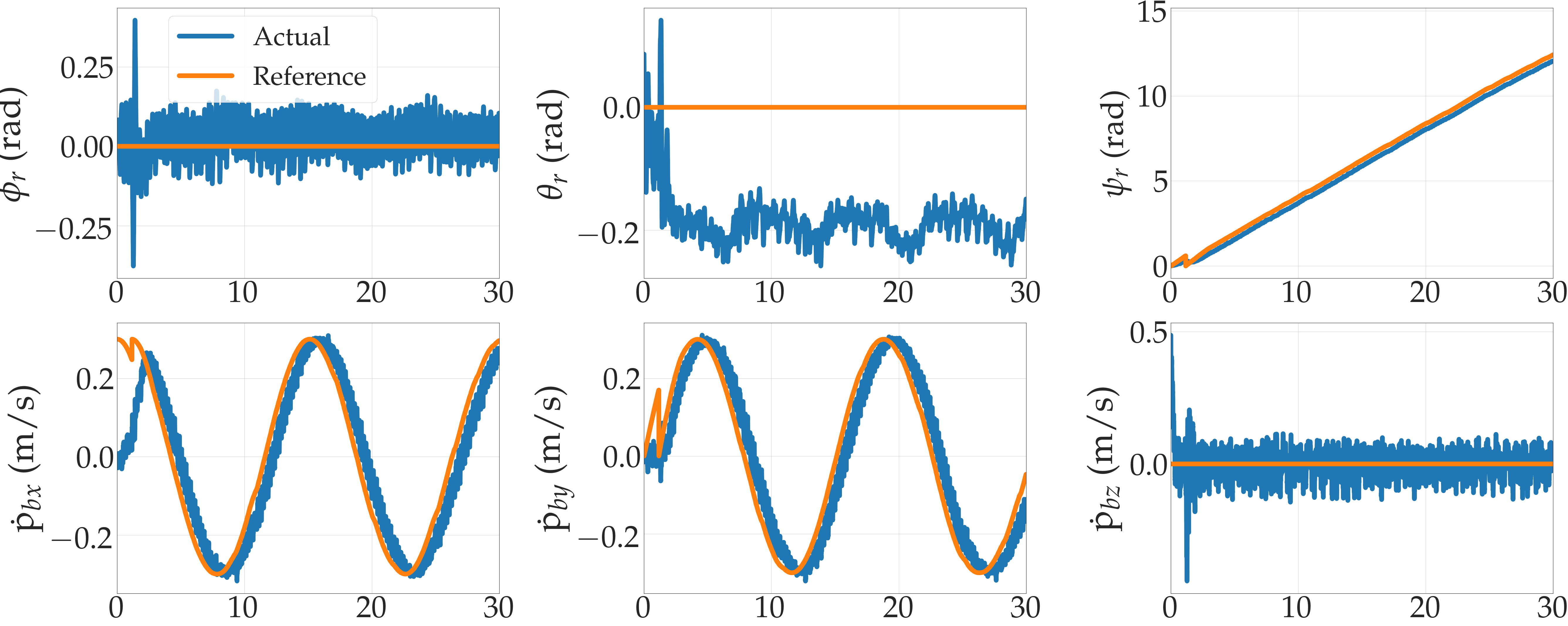}
    \caption{The simulation results of Mini Cheetah manipulating a ball to track a circle. The orange lines are the reference commands, and the blue lines are the actual values. The first row shows Euler angles (roll $\phi$, pitch $\theta$, and yaw $\psi$), and the second row shows robot velocities ($\dot x$, $\dot y$, $\dot z$).}
    \label{fig:ScenarioCircle}
\end{figure*}


Ball parameters are important factors affecting the stability and performance of the robot on the ball.
The lighter or the smaller the ball is, the more difficult it is to balance on it. Our controller can work with more extreme ball parameters.
Without changing the controller parameters, the minimum ball mass and radius for stabilization are 0.5kg and 0.5m, comparing with the 4kg and 0.8m for the baseline controller.

One of the main assumptions that we have made in our control design and simulation is the ball's rigidity.
For a more general case, such as stabilizing Mini Cheetah on a deformable object, e.g., a fitness ball, we need to take the deformation of that object into account during the dynamic modeling and control design.

%% file: conclusion.tex
\section{Conclusion}
\label{sec:conclusion}
In this paper, we have presented a novel control design for dynamic legged locomotion with an application to a quadrupedal robot manipulating a rigid ball. 
The control design for ball manipulation consists of a reaction-force-oriented model predictive controller and a foot placement controller, applied to an interaction model and integrated with a nominal WBIC. 
We numerically verified the control strategy with a variety of scenarios. The proposed controller allows a Mini Cheetah robot to manipulate a ball along straight/sinusoid/circular trajectories and outperforms a baseline controller. Experimental results are envisaged for the future. 

\section*{Acknowledgement}
The authors would like to thank Professor Sangbae Kim and the MIT Biomimetic Robotics Lab for providing the Mini Cheetah simulation software.